# Granular conditional entropy-based attribute reduction for partially labeled data with proxy labels

Can Gao[a,b], Jie Zhou[a,b,*], Duoqian Miao[c], Xiaodong Yue[d], Jun Wan[a,b]

[a]*College of Computer Science and Software Engineering, Shenzhen University*
*Shenzhen 518060, P.R. China*
[b]*SZU Branch, Shenzhen Institute of Artificial Intelligence and Robotics for Society*
*Shenzhen 518060, P.R. China*
[c]*Department of Computer Science and Technology, Tongji University*
*Shanghai 201804, China*
[d]*School of Computer Engineering and Science, Shanghai University*
*Shanghai 200444, China*

**Abstract**
Attribute reduction is one of the most important research topics in the theory of rough sets, and many rough sets-based attribute reduction methods have thus been presented. However, most of them are specifically designed for dealing with either labeled data or unlabeled data, while many real-world applications come in the form of partial supervision. In this paper, we propose a rough sets-based semi-supervised attribute reduction method for partially labeled data. Particularly, with the aid of prior class distribution information about data, we first develop a simple yet effective strategy to produce the proxy labels for unlabeled data. Then the concept of information granularity is integrated into the information-theoretic measure, based on which, a novel granular conditional entropy measure is proposed, and its monotonicity is proved in theory. Furthermore, a fast heuristic algorithm is provided to generate the optimal reduct of partially labeled data, which could accelerate the process of attribute reduction by removing irrelevant examples and excluding redundant attributes simultaneously. Extensive experiments conducted on UCI data sets demonstrate that the proposed semi-supervised attribute reduction method is promising and even compares favourably with the supervised methods on labeled data and unlabeled data with true labels in terms of classification performance.

## 1. Introduction

In many real-world applications, such as image classification, text mining, and gene analysis, the data to be processed is described by hundreds and thousands of attributes, which poses a substantial challenge for conventional data analysis [2]. Attribute reduction [17, 21] (a.k.a feature selection) has been proved to be an effective process of selecting the most informative attributes and removing the irrelevant or redundant attributes from data. Due to the merit of enhancing learning performance, increasing computational efficiency, improving interpretability, and alleviating over-fitting, attribute reduction has become an important pre-processing step in machine learning, pattern recognition, and data mining [1].

---

*Corresponding author at: College of Computer Science and Software Engineering, Shenzhen University, Shenzhen, 518060, P.R. China. Email addresses: **jie_jpu@163.com**.

[1] Our codes and experimental data are released at Mendeley Data http://dx.doi.org/10.17632/v3byhx2v8s.1 and Github https://github.com/Davidgaocan/SSL_FS.

The theory of rough sets [29] is a representative soft computing methodology for dealing with vague, uncertain, or imprecise data. Since the pioneering work of Pawlak [28], it has been witnessed rapid development in both theory and application [45, 47]. Attribute reduction [11, 16, 35] is one of the most important research issues in rough sets. The objectiveness of attribute reduction based on rough sets is to find an attribute subset that keeps the same discriminative ability as the original attribute set. In terms of the measure to evaluate the informativeness of attributes, attribute reduction based on rough sets can be roughly categorized into the positive region [9, 32], discernibility matrix [24, 43], and information-theoretic methods [6, 23, 30]. In positive region-based methods, all examples under the entire set of condition attributes are classified into positive and boundary examples, and the objectiveness of attribute reduct-ion is to search for an attribute subset that could hold the number of positive examples unchanged. Discernibility matrix-based methods first construct a matrix to describe the discernible attributes between each pair of examples, and the attribute subset that has a non-empty intersection with each non-empty element within the matrix is considered as a reduct. Information entropy is an efficient measure for uncertainty, and the information-theoretic-based methods aim to select a set of informative attributes that preserves the overall uncertain-ty of data as the original attribute set. Besides, to obtain the reduct with minimum cost/risk [39, 40, 41, 44], many optimization-based attribute reduction methods [5, 12, 31, 42] have also been proposed.

The methods mentioned above are often used to deal with either labeled data or unlabeled data. However, many real-world applications, such as web-page categorization, medical diagnosis, and defect detection [3, 49, 50], involve both labeled and unlabeled data. Therefore, semi-supervised attribute reduction based on rough sets is worthy of in-depth study. To tackle the data with partial supervision (referred to as partially labeled data hereafter), the concepts of semi-supervised discernibility matrix [22, 46] and discernibility pair [4] have been developed to yield the reduct of partially labeled data with categorical attributes since the discernibility matrix can be used to extract the discernible information of both labeled and unlabeled data. Instead of equivalence relation for categorical data, fuzzy dependency [10], neighbourhood approximate quality [19], neighbourhood decision error [20], neighbourhood granulation [15] have been provided to handle the partially labeled data with numerical attributes. Additionally, many other rough sets-based methods have been proposed for semi-supervised classification [13, 25, 26, 27, 34, 37] and semi-supervised clustering [18, 38].

The aforementioned works present some semi-supervised attribute reduction methods, but these methods have their limitations. On the one hand, to obtain the semi-supervised reduct, some of the existing methods use the carefully designed and complicated mechanism to generate the labels for unlabeled data, which severely limits their applicability to real-world tasks. On the other hand, the efficiency in search algorithm is also an important factor for attribute reduction method, while the process of generating the semi-supervised reduct in existing methods is still time-consuming. To tackle these limitations, in this paper, we propose an effective semi-supervised attribute reduction method for partially labeled data. The main contribution of this paper is threefold.

(1) To avoid the complex mechanism to annotate unlabeled data, a labelling strategy is designed, in which the class distribution information about the whole partially labeled data is considered as prior knowledge and is used along with the distribution of labeled data to determine the proxy label of unlabeled data. This strategy is very simple yet effective and has better adaptation for practical application.

(2) To better evaluate attributes, a novel information-theoretic measure is proposed for attribute reduction, which incorporates information granularity with conditional entropy. Furthermore, the monotonicity of the proposed measure is theoretically proved.

(3) To quicken the process of attribute reduction, we develop a strategy to accelerate the search algorithm by excluding unnecessary examples and filtering redundant attributes simultaneously. Moreover, extensive experiments are performed to verify the effectiveness of the proposed model, and very promising results are achieved.

The rest of the paper is organized as follows. Section 2 presents the preliminaries on rough sets and semi-supervised attribute reduction. Section 3 elaborates on the proposed semi-supervised attribute method for partially labeled data. Experimental analysis is conducted in Section 4. Finally, Section 5 concludes the paper and indicates the future work.

## 2. Preliminaries

This section will briefly review the basic concepts related to rough sets and semi-supervised learning. More details about these theories could refer to [3, 28, 29, 36, 50].

### 2.1. Rough sets

In rough sets, the data of interest is called an information system [29] and is denoted as $IS = (U, A)$, where $U$ is the set of examples, called the universe, and $A$ is the set of attributes to describe the examples. To be more specific, the information system is also called a decision information system or decision table if $A = C \cup D$, where $C$ is the set of condition attributes and $D$ is the decision attribute [29].

Given an attribute subset $B$ of $A$, the universe $U$ is partitioned into a family of equivalence classes $U/B$. An equivalence class containing $x$ is denoted as $[x]_B$ and is referred to as $B$-elementary granule [29]. Let $X$ be a subset of the universe $U$. Then, the lower and upper approximations of $X$ with respect to $B$ are defined as [29]:

$$\underline{B}(X) = \bigcup \{x \in U : [x]_B \subseteq X\},$$
$$\overline{B}(X) = \bigcup \{x \in U : [x]_B \cap X \neq \emptyset\}. \qquad (1)$$

The $B$-lower approximation of $X$ is the set of examples whose $B$-elementary granules belong to $X$, whereas the $B$-upper approximation of $X$ is the set of examples whose $B$-elementary granules have a non-empty intersection with $X$. $X$ is called a rough set with respect to $B$ if $\underline{B}(X) \neq \overline{B}(X)$; otherwise $X$ is a crisp set.

Let $IS = (U, A = C \cup D)$ be a decision table and $U/D = \{Y_1, Y_2, \ldots, Y_{|U/D|}\}$ be the partition induced by the decision attribute $D$ over $U$. Then, the positive, boundary, and negative regions of $D$ with respect to $C$ are defined as [29]:

$$POS_C(D) = \bigcup_{Y_i \in U/D} \underline{C}(Y_i)$$
$$BND_C(D) = \bigcup_{Y_i \in U/D} (\overline{C}(Y_i) - \underline{C}(Y_i)) \qquad (2)$$
$$NEG_C(D) = U - \bigcup_{Y_i \in U/D} \overline{C}(Y_i)$$

Let *MES* be a measure to quantify the correlation between the condition attributes and the decision attribute. Then, for an attribute subset $P$ of $C$, $P$ is a reduct of $C$ with respect to $D$ if and only if [29]:

**(I)** $MES_P(D) = MES_C(D)$, and

**(II)** $\forall a \in P \wedge P^* = P - \{a\}, \ MES_{P^*}(D) \neq MES_C(D)$.

The condition (I) is to guarantee the data after attribute reduction has the same descriptive ability as the original data, and the classification ability is thus preserved. While the condition (II) is to keep the attribute subset with the minimum redundancy. In other words, each attribute in the reduct is individually necessary. In rough sets, the measure *MES* could be positive region[29], information entropy[23], discernibility preservation [43], etc.

*2.2. Semi-supervised learning*

Semi-supervised learning is an efficient methodology for partially labeled data. Generally, partially labeled data $PS = (U = L \cup N, A = C \cup D)$ is a combination of two sets of examples: the labeled set $L = \{x_i, y_i\}_{i=1}^{l}$ and the unlabeled set $N = \{x_i, ?\}_{i=l+1}^{u=l+n}$, where $l$ is the number of labeled examples, $n$ is the number of unlabeled examples and $l \ll n$. In the context of semi-supervised learning, the label information of labeled data can be used to enhance the results of unsupervised clustering, called semi-supervised clustering [50]. Also, the geometric structure of unlabeled data can be captured to improve the performance of supervised method trained only on the labeled data, called semi-supervised attribute reduction, semi-supervised classification, or semi-supervised regression [36]. The detailed description of these methods could refer to [3, 49, 50]. In this paper, we only focus on semi-supervised attribute reduction.

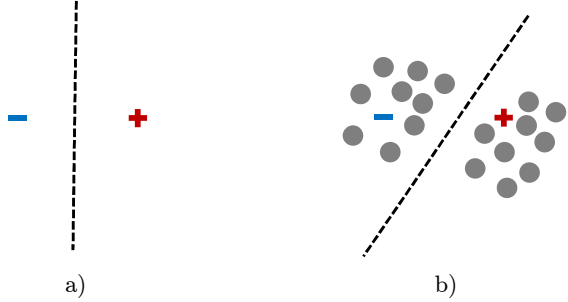

Figure 1. Semi-supervised learning. a) decision boundary using only labeled data; b) decision boundary using both labeled and unlabeled data

Semi-supervised attribute reduction is to employ a large amount of unlabeled data to aid the selection of informative attributes when the labeled data at hand are scarce. Like traditional supervised attribute reduction, semi-super-vised attribute reduction can be categorized into filter, wrapper, and embedded methods [33]. However, most of the existing methods are developed to cater for partially labeled data with numerical attributes. Little attention has paid to partially labeled data with categorical attributes.

## 3. Semi-supervised attribute reduction for partially labeled data

In this section, we first describe the strategy to generate proxy labels for unlabeled data. An improved information-theoretic measure is then developed, and a heuristic semi-supervised attribute reduction algorithm is proposed for partially labeled data with proxy labels.

### 3.1. Proxy label generation guided by prior knowledge

Traditional attribute reduction methods in rough sets are developed for labeled data or unlabeled data. When facing partially labeled data, attribute reduction operated on only labeled data may be insufficient since a large number of unlabeled data are available. While unsupervised attribute reduction performed on unlabeled data and labeled data without labels results in the waste of valuable label information. A promising way is to use both labeled and unlabeled data to carry out the process of attribute reduction. In this paper, we consider a strategy of annotating unlabeled data with proxy labels.

In practical semi-supervised application, there is domain-specific knowledge that can be utilized to facilitate the learning process. For example, in the detection of lung cancer, a large number of medical images can be easily collected in routine diagnosis, but only a few representative images may be labeled by medical experts since labelling all images is expensive and time-consuming. However, the occurrence probability of disease is generally known by domain experts in advance. Therefore, we could use this prior information to aid the learning of partially labeled data. More specifically, the distribution of different classes in practical tasks is regarded as the domain prior knowledge, while the distribution of initially labeled data in partially labeled data is used as the explicitly intrinsic information. Then the two information is integrated to determine the proxy labels of unlabeled data in partially labeled data.

Formally, we assume partially labeled data $PS = (U = L \cup N, A = C \cup D)$ has labeled data $L$ with $l$ examples and unlabeled data $N$ with $n$ examples, where $u = l + n$. Without loss of generality, we consider that the partially labeled data only have two classes, namely binary classification problem. Among the initially labeled data $L$, the sets of positive and negative examples are denoted as $L_{pos}$ and $L_{neg}$, respectively, and the ratio of positive examples over negative examples is denoted as $\gamma = |L_{pos}|/|L_{neg}|$. The prior probability of positive examples over all examples is denoted as $P_{pos}(U) = |U_{pos}|/|U|$.

Considering the prior information $P_{pos}(U)$ of partially labeled data and the class distribution of initially labeled data, the unlabeled examples in partially labeled data are determined to the proxy label $y_{proxy}$ by the following formula:

$$y_{proxy} = \begin{cases} y_{pos}, & \lambda \leq 0.5 \\ y_{neg}, & \lambda > 0.5 \end{cases} \qquad (3)$$

where $\lambda = P_{init}(\delta, \varepsilon) * P_{prior}(\varepsilon)$ and

$$P_{init}(\delta, \varepsilon) = \begin{cases} \gamma^{(1+e^{-\varepsilon \delta |L|})}, & |L| \leq \delta \\ 1, & |L| > \delta \end{cases} \qquad (4)$$

$$P_{prior}(\varepsilon) = \begin{cases} min(P_{pos}(U) * (1+\varepsilon)^{|U|}, 0.5), & P_{pos}(U) \leq 0.5 \\ 1 - min((1 - P_{pos}(U)) * (1+\varepsilon)^{|U|}, 0.5), & P_{pos}(U) > 0.5 \end{cases} \qquad (5)$$

Formally, the determination of proxy labels for unlabeled data involves two correlative parts. The first part is closely related to the distribution of initially labeled data. When the number of initially labeled data is small, the ratio $\gamma$ of positive examples over negative examples is of great influence on the determination of proxy labels. When $\gamma < 1$, it means that the number of positive labeled examples is smaller than that of negative labeled examples. In other words, the class distribution is unbalanced, which usually brings an adverse effect on the building of learning model. Therefore, the initial part $P_{init}$ tries to strengthen this unbalanced problem and make the labelling strategy assign positive proxy labels for unlabeled data. The smaller the number of initially labeled data, the greater the imbalance and the higher probability of labelling positive class for unlabeled data. And vice versa. On the case that the ratio $\gamma = 1$, the distribution of different classes is relatively balanced. The learning model will not suffer from class imbalance and the labelling strategy thus does not need to consider this part. However, the effect of the ratio $\gamma$ will gradually weaken as the number of initially labeled data increases since the adverse effect in class imbalance could partly remedy by enlarging the scale of examples. Therefore, we use a truncation function to suppress the effect of initially labeled data on the determination of proxy labels. Figure 2 shows the effect of the first part on the labelling strategy under different parameters.

The second part embodies the combined effect of prior knowledge and the scale of data. The prior probability is very useful information to conduct the assignment of proxy labels. When the prior probability of positive class is smaller than 0.5, initially labeled data will contain fewer positive examples. The labelling strategy should assign a positive proxy label to unlabeled data to enrich the set of positive examples. And vice versa. Further, the scale of data is also of great importance. The labelling strategy relies heavily on the prior probability when the scale of data is small. However, as the scale of data increases, the prior part $P_{prior}$ tends to 0.5. In other words, the proxy label can be arbitrarily assigned when the scale of data is very large and initially labeled data has a certain number of examples. Figure 3 demonstrates the effect of the second part on the labelling strategy under different parameters.

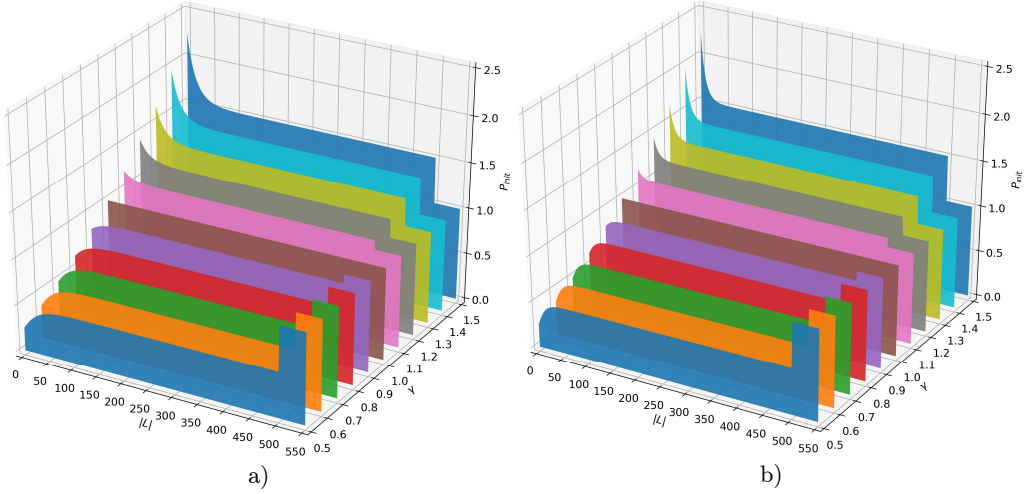

Figure 2. The effect of class distribution of initially labeled data on the labelling strategy. a) $P_{init}$ when $\delta = 500$ and $\varepsilon = 0.0001$; b) $P_{init}$ when $\delta = 500$ and $\varepsilon = 0.0002$

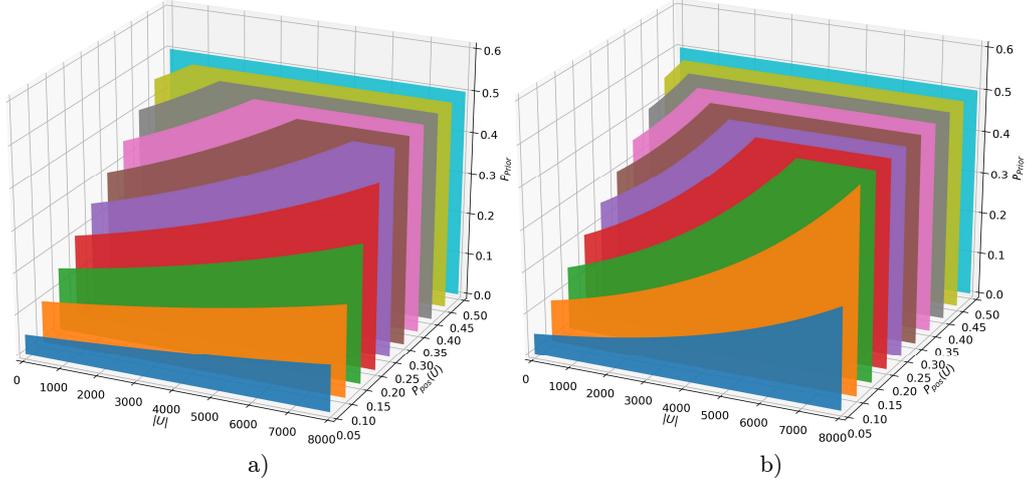

Figure 3. The effect of prior knowledge on the labelling strategy. a) $P_{prior}$ when $\varepsilon = 0.0001$; b) $P_{prior}$ when $\varepsilon = 0.0002$

The labelling strategy concerns the scale of data $|U|$, the number of initially labeled data $|L|$, the prior probability $P_{pos}(U)$, the class ratio $\gamma$, the boosting factor $\varepsilon$, and the truncation threshold $\delta$. As a matter of the fact, there are only two parameters $\varepsilon$ and $\delta$ that need to be set because the other parameters are automatically determined when a partially labeled data and prior knowledge are given. In the following, the two parameters are empirically set to $\varepsilon = 0.0002$ and $\delta = 500$, respectively, since they could trade off the effect of data scale, prior probability, and initially labeled data on the labelling strategy.

*3.2. Semi-supervised attribute reduction based on granular conditional entropy*

Information entropy is an efficient measure for uncertainty and is thus often used to estimate the correlation or redundancy between attributes. In this paper, we propose granular conditional entropy to evaluate the importance of attributes in partially labeled data with proxy labels. Formally, the partially labeled data after adopting the proposed labelling strategy is denoted as $PS = (U = L \cup N', A = C \cup D)$.

**Definition 1.** Let $PS = (U = L \cup N', A = C \cup D)$ be a partially labeled data with proxy labels and $U/B = \{X_1, X_2, ..., X_{|U/B|}\}$ be the partition induced by the condition attribute subset $B \subseteq C$. Then, the entropy of $B$ over $U$ is defined as [30]:

$$H(B) = -\sum_{i=1}^{|U/B|} P(X_i) \log P(X_i), \quad (6)$$

where $P(X_i) = |X_i|/|U|$ and $|\cdot|$ denotes the cardinality of a finite set.

**Definition 2.** Let $PS = (U = L \cup N', A = C \cup D)$ be a partially labeled data with proxy labels, $U/B = \{X_1, X_2, ..., X_{|U/B|}\}$ and $U/D = \{Y_1, Y_2, ..., Y_{|U/D|}\}$ be the partitions induced by the condition attribute subset $B \subseteq C$ and the decision attribute $D$, respetively. Then, the conditional entropy of $D$ given $B$ is defined as [30]:

$$H(D|B) = -\sum_{i=1}^{|U/B|} \sum_{j=1}^{|U/D|} P(X_i, Y_j) \log P(Y_j|X_i), \quad (7)$$

where $P(X_i, Y_j) = P(Y_j|X_i)/P(X_i)$ and $P(Y_j|X_i) = |X_i \cap Y_j|/|X_i|$.

**Definition 3.** Let $PS = (U = L \cup N', A = C \cup D)$ be a partially labeled data with proxy labels and $U/B = \{X_1, X_2, ..., X_{|U/B|}\}$ be the partition induced by the condition attribute subset $B \subseteq C$. Then, the granularity of $B$ over $U$ is defined as [14]:

$$G(B) = -\sum_{i=1}^{|U/B|} P(X_i)^2. \quad (8)$$

**Definition 4.** Let $PS = (U = L \cup N', A = C \cup D)$ be a partially labeled data with proxy labels, $U/B = \{X_1, X_2, ..., X_{|U/B|}\}$ and $U/D = \{Y_1, Y_2, ..., Y_{|U/D|}\}$ be the partitions induced by the condition attribute subset $B \subseteq C$ and the decision attribute $D$, respetively. Then, the granular conditional entropy of $D$ given $B$ is defined as:

$$GH(D|B) = -\sum_{i=1}^{|U/B|} P(X_i)^2 \sum_{j=1}^{|U/D|} P(Y_j|X_i) \log P(Y_j|X_i). \quad (9)$$

Compared with the measure of conditional entropy, granular conditional entropy incorporates the granularity of attributes, while the granularity information essentially reflects the discriminability of attributes. The finer the granularity of attributes, the stronger the discriminating power. Therefore, the reduct with more informative attributes could be obtained by the proposed granular conditional entropy.

**Proposition 1:** Let $PS = (U = L \cup N', A = C \cup D)$ be a partially labeled data with proxy labels. Then, for any attribute subset $B \subseteq C$, $0 \leq GH(D|B) \leq \log|U|$.

**Proof.** Granular conditional entropy attains the minima when the decision attribute $D$ is completely dependent on the condition attribute subset $B$. In other words, each equivalence class induced by the condition attribute subset $B$ has only one decision, while the conditional probability of $D$ given $B$ is 0 and the overall granular conditional entropy is thus minimized to 0. The granular conditional entropy arrives at the maxima when the decision attribute $D$ is conditionally independent of the condition attribute subset $B$, namely $GH(D|B) = H(D)$. While $H(D)$ achieves the maximum value $\log|U|$ when the probability distribution is uniform. Thus $GH(D|B) \leq \log|U|$. The proposition is proved.

**Proposition 2:** Let $PS = (U = L \cup N', A = C \cup D)$ be a partially labeled data with proxy labels, and $P, Q$ be the subsets of $C$. If $P \subset Q$, then $GH(D|P) \geq GH(D|Q)$.

**Proof.** Without loss of generality, assume $Q = P \cup \{a\}$ and only the equivalence class $X_{ij}$ under $P$ is divided into the equivalence classes $X_i$ and $X_j$ under $Q$ after adding the attribute $a$. Namely, $U/P = \{X_1, X_2, \ldots, X_{ij}, \ldots, X_n\}$ and $U/Q = \{X_1, X_2, \ldots, X_i, X_j, \ldots, X_n\}$. The proof is presented in the appendix.

Proposition 2 guarantees that granular conditional entropy is monotonically decreasing as adding attributes. Thus, it can be considered as a measure for attribute reduction.

**Definition 5.** Let $PS = (U = L \cup N', A = C \cup D)$ be a partially labeled data with proxy labels, and let $P \subset C$. Then, for a condition attribute $a \in (C - P)$, the relative significance for $D$ given $P$ is defined as:

$$Sig(a, P, D) = GH(D|P) - GH(D|(P \cup \{a\})). \tag{10}$$

**Definition 6.** Let $PS = (U = L \cup N', A = C \cup D)$ be a partially labeled data with proxy labels. Then, for an attribute subset $P$ of $C$, $P$ is a reduct of $C$ with respect to $D$ if and only if:

**(I)** $GH(D|P) = GH(D|C)$, and

**(II)** $\forall a \in P \land P^* = P - \{a\}$, $GH(D|P^*) \neq GH(D|C)$.

Attribute reduction is highly related to the measure for evaluating the significance of attributes and also involves the strategy of finding the reduct. It is well-known that finding the minimum reduct or all reducts is NP-hard so that the heuristic method is preferred. While the existing heuristic methods can be further divided into the strategies of forward adding, backward deleting, and bi-directional adding-deleting. Considering the efficiency, we use the strategy of forward adding to conduct the process of attribute reduction for partially labeled data.

In terms of the characteristics of the proposed granular conditional entropy, we further design two tactics to accelerate the process of attribute reduction. On the one hand, granular conditional entropy is monotonically decreasing as adding attributes to the reduct. During the process of attribute reduction, if a condition equivalence class has granular conditional entropy 0 at one stage, its granular conditional entropy is always 0 in the following stages. Therefore, this kind of examples in the condition equivalence class can be removed without further consideration. On the other hand, to find the informative attributes for the reduct, the attribute reduction algorithm will examine the relative significance of each candidate attribute and select the optimal attributes to the reduct. Considering that attributes are correlated with each other, and there may be some attributes whose granular conditional entropy with respect to the selected optimal attributes is 0, which means the attributes are redundant to the decision attribute given the selected attributes. Thus, this kind of attributes can also be excluded from the list of candidate attributes for the reduct. The overall granular conditional entropy-based attribute reduction algorithm embedded with acceleration strategy can be depicted by Algorithm 1.

---

**Algorithm 1** An accelerated semi-supervised attribute reduction algorithm based on granular conditional entropy

---

**Input:**
　A partially labeled data $PS = (U = L \cup N, A = C \cup D)$, the prior probability of positive class $P_{pos}(U)$, and the threshold parameters $\delta$ and $\varepsilon$.
**Output:**
　An optimal semi-supervised reduct $RED$;
1: Compute the class ratio $\gamma$ within the initially labeled data $L$;
2: Determine the proxy labels of unlabeld data $N$ by the prior probability $P_{pos}(U)$, the class ratio $\gamma$, and the threshold parameters $\delta$ and $\varepsilon$; //refers to Formula (3)
3: Compute the overall granular conditional entropy $GH(D|C)$;
4: Evaluate each attribute by the granular conditional entropy $GH(D|\{a_i\})$, and add the attribute $a_{opt} = argmin_{a_i \in C}\{GH(D|\{a_i\})\}$ to $RED$;
5: **While** $GH(D|RED) \neq GH(D|C)$ **Do**
6: 　Compute the relative significance of each attriubte $a_i$ for $D$ given $RED$ and the granular conditional entropy $GH(\{a_i\}|RED)$;
7: 　Select an attribute $a_{opt}$ whose significance is maximal and remove the attributes whose $GH(\{a_i\}|RED)$ is 0; //Acceleration in attribute
8: 　$RED \leftarrow RED \cup \{a_{opt}\}$ and remove the examples whose granular conditional entropy under $RED$ is 0; //Acceleration in example
9: **End While**
10: **Return** The semi-supervised reduct $RED$.

---

The algorithm first label all unlabeled examples with the proxy labels determined by the prior probability and the class distribution of initially labeled data (line 1 and line 2). The overall granular conditional entropy under all conditional attributes is then computed and is considered as the stopping condi-tion for the algorithm. In the first round of attribute selection, the algorithm evaluates each attribute only by its granular conditional entropy with respect to the decision attribute and the attribute whose granular conditional entropy is minimal is selected as the optimal attribute. In the following rounds of attribute selection, the algorithm iteratively adds the optimal attributes with the maximum significance into the reduct until the granular conditional entropy of the selected attributes reaches the stopping value (line 5 to line 9). Wherein, two acceleration strategies are embedded into the algorithm, thus resulting in higher efficiency for attribute reduction.

The main cost of Algorithm 1 lies in the iterative selection of the optimal attributes. Assume that a partially labeled data has $|U|$ examples described by $|C|$ attributes. In each iteration, the time cost for determining an optimal attribute is $O(|C||U|^2)$. In the worst-case, the algorithm is terminated after $|C|$ rounds of selection. Therefore, the time cost for computing an optimal reduct of a given partially labeled data is at most $O(|C|^2|U|^2)$ and the total space cost is at most $O(|C||U|)$. Considering the acceleration strategy, the overall cost of Algorithm 1 is much lower in time and space.

## 4. Empirical analysis

In this section, we first verify the effectiveness of the proposed method for semi-supervised attribute reduction. Then we compare the proposed method with other classic methods in terms of classification accuracy. All experiments were carried out on a computer with Windows 10 operating system, Intel Xeon (R) CPU E5-2650 v4@2.20 GHz processor, and 128 GB Memory.

### 4.1. Investigated data sets and experiment design

Twelve UCI data sets[1] are used in the experiments, and the details are shown in Table 1. Note that some of the data sets are multi-classification tasks. The criterion of "1-vs-all" is employed to convert them into binary classification task. Specifically, the class that has the highest probability is considered as positive class and the remaining classes are grouped into negative class.

Table 1: The experimental data sets

| Data set | $|U|$ | $|C|$ | $|U/D|$ | $(P_{pos}(U), P_{neg}(U))$ |
|---|---|---|---|---|
| cardiotocography-FHR pattern(cardio) | 2126 | 21(21) | 10 | (0.2723, 0.7277) |
| frogs calls-species(frog) | 7195 | 22(22) | 8 | (0.5768, 0.4232) |
| gesture-phase-a3va3(gesture1) | 1830 | 32(32) | 5 | (0.3595, 0.6405) |
| gesture-phase-b1va3(gesture2) | 1069 | 32(32) | 5 | (0.3854, 0.6146) |
| kdd-synthetic-control(kdd) | 600 | 60(0) | 6 | (0.1667, 0.8333) |
| kr-vs-kp(krvskp) | 3196 | 36(0) | 2 | (0.5222, 0.4778) |
| landsat(landsat) | 6435 | 36(0) | 6 | (0.2382, 0.7618) |
| libras movement(libras) | 360 | 90(90) | 15 | (0.0667, 0.9333) |
| musk2(musk) | 6598 | 166(0) | 2 | (0.8458, 0.1542) |
| spambase(spam) | 4601 | 57(57) | 2 | (0.6060, 0.3940) |
| vehicle(vehicle) | 846 | 18(18) | 4 | (0.2577, 0.7423) |
| wine(wine) | 178 | 13(13) | 3 | (0.3989, 0.6011) |
| Avg. | 2919.50 | 48.58(23.75) | 5.67 | (0.3914, 0.6086) |

---

1. http://archive.ics.uci.edu/ml/index.php

In Table 1, the number of examples is shown in the second column. The number of condition attributes is presented in the third column, where the number of numerical attributes is also listed in the brackets. The number of classes in the original data set is given in the fourth column. While the last column denotes the class distribution after applying the "1-vs-all" criterion, which is considered as the prior knowledge to guide the determination of proxy labels.

In the experiments, each numerical attribute is discretized into categorical one using the technique of equal frequency binning with three bins. To fully examine the semi-supervised reduct of partially labeled data, the experiments were performed on different label rates $\alpha \in [0.01, 0.3]$. Under a given label rate $\alpha$, each data set is first partitioned into a set of labeled examples $L$ and a set of unlabeled data $N$. Then, the prior class probability and the class ratio of initially labeled data $L$ are used to determine the proxy labels of unlabeled examples $N$. To deeply gain insight the labelling strategy, the experiments were further performed on the data partition with different positive ratios. More specifically, under a given label rate $\alpha$, the positive ratio $\beta$ of initially labeled data varies from 0.5 to 1.5. For example, given a partially labeled data with 1000 examples, the prior probability $P_{pos}(U) = 0.5$ and a labeled rate $\alpha = 10\%$, a labeled set $L$ with 25 positive examples ($|L_{pos}| = \beta * P_{pos}(U) * \alpha * |U|$) and 75 negative examples ($|L_{neg}| = \alpha * |U| - |L_{pos}|$) is randomly generated when the positive ratio $\beta = 0.5$, and then the remaining 900 examples is grouped into a set of unlabeled examples. To guarantee the effectiveness of experimental results, we repeated the data partition for 10 times at each pair of label rate $\alpha$ and positive ratio $\beta$, and the performance is finally averaged.

*4.2. Attribute reduction for partially labeled data*

To test the effectiveness of the proposed attribute reduction algorithm for partially labeled data, we conducted an experiment on all examples of each data set under a label rate $\alpha = 10\%$, but the positive ratio $\beta$ varies from 0.5 to 1.5. The reduct information is shown in Table 2.

Table 2: The results of attribute reduction on the selected data sets (label rate $\alpha = 10\%$)

| Data set | Raw | Only labeled data | | | Ours | | | Ground-truth | Approximate rate |
|---|---|---|---|---|---|---|---|---|---|
| | | Min | Max | Avg. | Min | Max | Avg. | | |
| cardio | 21 | 6 | 11 | 8.04 | 17 | 20 | 19.19 | 17 | 0.89 |
| frog | 22 | 7 | 11 | 8.85 | 21 | 22 | 21.32 | 14 | 0.66 |
| gesture1 | 32 | 7 | 13 | 9.21 | 19 | 27 | 22.35 | 21 | 0.94 |
| gesture2 | 32 | 5 | 8 | 6.33 | 12 | 21 | 17.05 | 16 | 0.94 |
| kdd | 60 | 2 | 4 | 2.55 | 8 | 10 | 8.81 | 5 | 0.57 |
| krvskp | 36 | 10 | 18 | 13.59 | 29 | 33 | 31.49 | 29 | 0.92 |
| landsat | 36 | 7 | 18 | 12.18 | 36 | 36 | 36.00 | 29 | 0.81 |
| libras | 90 | 2 | 3 | 2.25 | 10 | 17 | 12.73 | 5 | 0.39 |
| musk | 166 | 2 | 14 | 8.07 | 37 | 72 | 57.86 | 24 | 0.41 |
| spam | 57 | 7 | 18 | 12.9 | 33 | 48 | 42.66 | 38 | 0.89 |
| vehicle | 18 | 3 | 7 | 4.90 | 12 | 16 | 14.09 | 10 | 0.71 |
| wine | 13 | 1 | 3 | 1.89 | 6 | 10 | 7.51 | 5 | 0.67 |
| Avg. | 48.58 | 4.91 | 10.67 | 7.56 | 20.00 | 27.67 | 24.26 | 17.75 | 0.73 |

In the table, the statistical information, including the minimum, the maximum, and the average numbers of attributes in the obtained reducts across different positive ratios, are listed in the third to eighth columns. Besides, we also record the ground-truth reduct information for comparison, namely the optimal reduct of data set under a label rate $\alpha = 100\%$. The last column "Approximate rate" indicates the similarity between the semi-supervised reduct and the ground-truth reduct, which can be computed by the value of "gound-truth" over that of "Avg." in the proposed method.

By observing the experimental results, we find that, on the one hand, the class distribution of initially labeled data has a great influence on the obtained reduct of partially labeled data. The more balanced the class distribution, the fewer attributes the reduct will have. On the other hand, the completely irrelevant attributes are always excluded from the obtained reducts, whereas different weakly relevant attributes will be removed from the reducts when different labeled examples are available. The reducts induced only from labeled data seems to attain the best attribute reduction rate, but they could only discern the labeled data rather than overall partially labeled data and these reducts are thus not good enough to classification, which will be confirmed by the following experiments. To discern both labeled and unlabeled data, the semi-supervised method has to select more informative attributes so that more discriminative ability is preserved. On all selected data sets, the proposed method achieves a reduction rate of 50.08% over raw data and an approximate rate of 73.18% with respect to ground-truth. It is worth mentioning that, on data sets "cardio", "gesture1", "gesture2", "krvskp", and "spam", the minimum number of attributes in the semi-supervised reduct is equal or even smaller than that of ground-truth, which further validates the potential of the proposed attribute reduction method for partially labeled data.

*4.3. The effectiveness of the proposed method*

To evaluate the quality of the reducts obtained by the proposed method, we further conducted performance experiments under a label rate $\alpha = 10\%$. Specifically, a supervised or semi-supervised reduct is first generated under the given label rate $\alpha = 10\%$ and a specific positive ratio $\beta \in [0.5, 1.5]$. The redundant attributes that are not contained in the obtained reduct are then removed from each data set, and the 10-fold cross-validation is performed on the reduced data set to realize performance evaluation. In the experiments, the classifiers of $k$-Nearest Neighbors with $k = 3$ and SVM with radial basis function are utilized, and the results are shown in Tables 3 and 4, respectively. Note that we shuffled each data set 10 times and performed 10-fold cross-validation on the shuffled data in order to avoid the impact of the order of samples on performance.

In Tables 3 and 4, we report the average performance over 10 run times of 10-fold cross-validation. At each positive ratio $\beta$, the columns "initial" and "final" denote the performance of the reduct obtained from initially labeled data and that of the semi-supervised reduct further refined by unlabeled data with proxy labels. The highest performance across different positive ratios is boldfaced and the average performance over all data sets is shown in the last row "Avg. ".

Table 3: The performance of the proposed method under different positive ratios (KNN with a label rate $\alpha = 10\%$)

| Data set | 0.5 | | 0.6 | | 0.7 | | 0.8 | | 0.9 | | 1.0 | | 1.1 | | 1.2 | | 1.3 | | 1.4 | | 1.5 | |
|---|---|---|---|---|---|---|---|---|---|---|---|---|---|---|---|---|---|---|---|---|---|---|
| | initial | final | initial | final | initial | final | initial | final | initial | final | initial | final | initial | final | initial | final | initial | final | initial | final | initial | final |
| cardio | 0.8557 | 0.8861 | 0.8539 | **0.8876** | 0.8591 | 0.8874 | 0.8601 | 0.8863 | 0.8679 | 0.8863 | 0.8622 | 0.8871 | 0.8615 | 0.8860 | 0.8637 | 0.8869 | 0.8674 | 0.8868 | 0.8660 | 0.8867 | 0.8611 | 0.8869 |
| frog | 0.9679 | 0.9862 | 0.9696 | **0.9863** | 0.9706 | 0.9863 | 0.9707 | 0.9861 | 0.9696 | 0.9857 | 0.9681 | 0.9858 | 0.9667 | 0.9859 | 0.9645 | 0.9858 | 0.9654 | 0.9858 | 0.9663 | 0.9858 | 0.9645 | 0.9858 |
| gesture1 | 0.7899 | 0.8528 | 0.7996 | 0.8570 | 0.8035 | 0.8563 | 0.8173 | 0.8560 | 0.8066 | 0.8507 | 0.8086 | 0.8512 | 0.8015 | 0.8528 | 0.8045 | **0.8610** | 0.8067 | 0.8524 | 0.8117 | 0.8592 | 0.8076 | 0.8570 |
| gesture2 | 0.7009 | **0.7845** | 0.7033 | 0.7711 | 0.6900 | 0.7712 | 0.6985 | 0.7780 | 0.6966 | 0.7662 | 0.6942 | 0.7670 | 0.6955 | 0.7734 | 0.6952 | 0.7659 | 0.6976 | 0.7723 | 0.6931 | 0.7455 | 0.6919 | 0.7575 |
| kdd | 0.9048 | 0.9815 | 0.9203 | 0.9783 | 0.9412 | 0.9760 | 0.9427 | 0.9752 | 0.9443 | 0.9808 | 0.9230 | 0.9783 | 0.9212 | 0.9743 | 0.9385 | **0.9833** | 0.9357 | 0.9782 | 0.9305 | 0.9817 | 0.9325 | 0.9793 |
| krvskp | 0.9344 | 0.9473 | 0.9308 | 0.9479 | 0.9299 | 0.9473 | 0.9329 | 0.9464 | 0.9298 | 0.9460 | 0.9355 | 0.9449 | 0.9311 | **0.9493** | 0.9349 | 0.9489 | 0.9333 | 0.9493 | 0.9400 | 0.9479 | 0.9283 | 0.9491 |
| landsat | 0.9501 | 0.9769 | 0.9571 | 0.9769 | 0.9582 | 0.9769 | 0.9557 | 0.9766 | 0.9604 | 0.9770 | 0.9607 | 0.9771 | 0.9605 | **0.9772** | 0.9609 | 0.9770 | 0.9594 | 0.9770 | 0.9623 | 0.9771 | 0.9633 | 0.9767 |
| libras | 0.8603 | 0.9672 | 0.8997 | 0.9664 | 0.8475 | 0.9608 | 0.8658 | 0.9664 | 0.8594 | 0.9611 | 0.8617 | 0.9631 | 0.8800 | **0.9714** | 0.8875 | 0.9625 | 0.8903 | 0.9617 | 0.8667 | 0.9614 | 0.8389 | 0.9683 |
| musk | 0.9179 | 0.9453 | 0.9239 | 0.9492 | 0.9228 | 0.9509 | 0.9249 | 0.9490 | 0.9174 | **0.9517** | 0.9178 | 0.9511 | 0.9050 | 0.9508 | 0.7661 | 0.9502 | 0.7729 | 0.9513 | 0.7956 | 0.9509 | 0.8063 | 0.9513 |
| spam | 0.8911 | 0.9249 | 0.8854 | 0.9228 | 0.8894 | 0.9189 | 0.8940 | 0.9206 | 0.8853 | 0.9249 | 0.8911 | 0.9229 | 0.8876 | 0.9218 | 0.8860 | 0.9227 | 0.8782 | 0.9192 | 0.8835 | 0.9221 | 0.8633 | **0.9251** |
| vehicle | 0.9123 | 0.9490 | 0.9019 | 0.9507 | 0.8903 | 0.9476 | 0.9126 | 0.9491 | 0.9115 | 0.9450 | 0.9079 | 0.9455 | 0.9214 | 0.9472 | 0.9211 | **0.9512** | 0.9123 | 0.9473 | 0.9119 | 0.9475 | 0.9084 | 0.9455 |
| wine | 0.7551 | 0.9029 | 0.7999 | 0.9104 | 0.8266 | 0.9067 | 0.8146 | 0.9234 | 0.7697 | 0.9133 | 0.8425 | 0.9178 | 0.7845 | **0.9239** | 0.7616 | 0.9166 | 0.8016 | 0.9234 | 0.8249 | 0.9126 | 0.8094 | 0.8729 |
| Avg. | 0.8700 | 0.9254 | 0.8788 | 0.9254 | 0.8774 | 0.9239 | 0.8825 | 0.9261 | 0.8766 | 0.9241 | 0.8811 | 0.9243 | 0.8764 | **0.9262** | 0.8654 | 0.9260 | 0.8684 | 0.9254 | 0.8710 | 0.9232 | 0.8646 | 0.9213 |

Table 4: The performance of the proposed method under different positive ratios (SVM with a label rate $\alpha = 10\%$)

| Data set | 0.5 | | 0.6 | | 0.7 | | 0.8 | | 0.9 | | 1.0 | | 1.1 | | 1.2 | | 1.3 | | 1.4 | | 1.5 | |
|---|---|---|---|---|---|---|---|---|---|---|---|---|---|---|---|---|---|---|---|---|---|---|
| | initial | final | initial | final | initial | final | initial | final | initial | final | initial | final | initial | final | initial | final | initial | final | initial | final | initial | final |
| cardio | 0.8306 | 0.8697 | 0.8374 | 0.8744 | 0.8438 | 0.8702 | 0.8475 | 0.8743 | 0.8502 | 0.8738 | 0.8465 | 0.8721 | 0.8444 | 0.8672 | 0.8490 | 0.8744 | 0.8455 | **0.8753** | 0.8421 | 0.8717 | 0.8398 | 0.8676 |
| frog | 0.9313 | 0.9486 | 0.9314 | **0.9486** | 0.9368 | 0.9485 | 0.9349 | 0.9484 | 0.9368 | 0.9482 | 0.9310 | 0.9482 | 0.9344 | 0.9482 | 0.9315 | 0.9482 | 0.9279 | 0.9482 | 0.9302 | 0.9482 | 0.9329 | 0.9482 |
| gesture1 | 0.7322 | 0.7570 | 0.7367 | 0.7603 | 0.7314 | 0.7589 | 0.7308 | 0.7570 | 0.7374 | 0.7563 | 0.7365 | 0.7480 | 0.7328 | 0.7580 | 0.7411 | 0.7548 | 0.7290 | 0.7541 | 0.7478 | **0.7675** | 0.7433 | 0.7674 |
| gesture2 | 0.6458 | 0.6522 | 0.6355 | 0.6504 | 0.6373 | 0.6542 | 0.6351 | 0.6492 | 0.6473 | 0.6488 | 0.6261 | 0.6500 | 0.6481 | 0.6622 | 0.6475 | 0.6549 | 0.6460 | **0.6625** | 0.6314 | 0.6485 | 0.6387 | 0.6504 |
| kdd | 0.9262 | 0.9803 | 0.9167 | 0.9765 | 0.9345 | 0.9772 | 0.9530 | 0.9813 | 0.9392 | **0.9833** | 0.9433 | 0.9755 | 0.9342 | 0.9712 | 0.9520 | 0.9818 | 0.9432 | 0.9753 | 0.9397 | 0.9823 | 0.9420 | 0.9780 |
| krvskp | 0.9465 | 0.9587 | 0.9445 | 0.9588 | 0.9479 | 0.9585 | 0.9483 | 0.9586 | 0.9467 | 0.9585 | 0.9472 | 0.9589 | 0.9462 | 0.9589 | 0.9485 | 0.9589 | 0.9450 | **0.9592** | 0.9473 | 0.9588 | 0.9465 | 0.9591 |
| landsat | 0.9299 | 0.9663 | 0.9334 | 0.9663 | 0.9407 | 0.9663 | 0.9369 | 0.9663 | 0.9410 | 0.9663 | 0.9397 | 0.9663 | 0.9372 | 0.9663 | 0.9374 | 0.9663 | 0.9357 | 0.9663 | 0.9392 | 0.9663 | 0.9435 | 0.9663 |
| libras | 0.9333 | 0.9547 | 0.9333 | 0.9508 | 0.9333 | 0.9494 | 0.9333 | 0.9489 | 0.9331 | 0.9514 | 0.9333 | **0.9597** | 0.9389 | 0.9589 | 0.9333 | 0.9533 | 0.9333 | 0.9531 | 0.9344 | 0.9467 | 0.9333 | 0.9589 |
| musk | 0.8572 | 0.9007 | 0.8591 | 0.9074 | 0.8626 | 0.9134 | 0.8547 | 0.9092 | 0.8607 | 0.9157 | 0.8548 | 0.9149 | 0.8528 | 0.9141 | 0.8459 | **0.9176** | 0.8459 | 0.9161 | 0.8459 | 0.9159 | 0.8459 | 0.9122 |
| spam | 0.8918 | **0.9403** | 0.8912 | 0.9380 | 0.8952 | 0.9375 | 0.8988 | 0.9388 | 0.8942 | 0.9395 | 0.8937 | 0.9389 | 0.8936 | 0.9401 | 0.8901 | 0.9394 | 0.8850 | 0.9395 | 0.8823 | 0.9400 | 0.8673 | 0.9393 |
| vehicle | 0.8561 | 0.9119 | 0.8519 | 0.9122 | 0.8266 | 0.9101 | 0.8353 | **0.9216** | 0.8476 | 0.9091 | 0.8488 | 0.9059 | 0.8531 | 0.9166 | 0.8544 | 0.9197 | 0.8454 | 0.9082 | 0.8391 | 0.9098 | 0.8325 | 0.8983 |
| wine | 0.8461 | 0.9125 | 0.8388 | 0.9170 | 0.8068 | 0.9276 | 0.8406 | 0.9315 | 0.8971 | 0.9193 | 0.8875 | 0.9249 | 0.8478 | 0.9292 | 0.8613 | **0.9327** | 0.8380 | 0.9276 | 0.8470 | 0.9125 | 0.7989 | 0.8622 |
| Avg. | 0.8606 | 0.8961 | 0.8592 | 0.8967 | 0.8581 | 0.8976 | 0.8624 | 0.8988 | 0.8693 | 0.8975 | 0.8657 | 0.8969 | 0.8636 | 0.8992 | 0.8660 | **0.9002** | 0.8600 | 0.8988 | 0.8605 | 0.8974 | 0.8554 | 0.8923 |

From Tables 3 and 4, we can see that the quality of attribute reduction is significantly improved by unlabeled data. Since the scarcity of label information in partially labeled data, the attribute evaluation only on labeled data may not really reflect the importance of attributes such that the supervised attribute reduction method generates low-quality reducts with few attributes and the mediocre performance is consequently obtained. Essentially, the semi-supervised reduct takes into consideration both labeled and unlabeled data. Moreover, the attribute selection is guided by the proxy labels of unlabeled data, which is jointly determined by prior knowledge and the class information of initially labeled data. As a result, the selected attributes are more representa-tive and informative and higher performance is achieved by the resulting semi-supervised reduct. On each data set, the supervised reduct gains different performance when varying the positive ratio from 0.5 to 1.5, while the performance of the obtained semi-supervised reduct is relatively stable and high. One possible reason for this might be that the balance of class distribution has great effect on the performance, and the semi-supervised reduct could weaken this adverse effect by utilizing the proxy labels of unlabeled data. By averaging all results across different data sets, the proposed method using KNN and SVM achieves a maximum improvement of 7.01% ($\beta = 1.2$) and 4.61% ($\beta = 0.7$), over the supervised method, respectively. Interestingly, the proposed method reaches the highest performance when the positive ratios $\beta = 1.1$ (KNN) and $\beta = 1.2$ (SVM), respectively, at which the class ratio on initially labeled data is close to 0.5 ($\beta * P_{pos}(U)$). In other words, the proposed method is likely to gain the highest performance when the class distribution of initially labeled data is balanced. These results clearly indicate that the proposed method is effective and could benefit from the proxy labels of unlabeled data to improve the quality of attribute reduction.

To further verify the effectiveness of the proposed method, it is compared to other attribute reduction methods, including supervised Fisher score (FS) [7], unsupervised Laplacian score (LS) [8], the proposed granular conditional entropy with only labeled data (GCE-L), the proposed granular conditional entropy with all data labeled (GT), and the raw data without attribute reduction (Raw). The settings for all selected methods are shown in Table 5. The experiments were performed under different label rates, and the results are shown in Figures 4 and 5. Note that the performance of the selected GCE-L, FS, and our method is averaged across different positive ratios.

As shown in Figures 4 and 5, it is obvious that the proposed method significantly outperforms the supervised methods with only initially labeled data on almost all data sets. Although Fisher score (FS) and granular conditional entropy (GCE-L) are effective measures for attribute reduction, the important attributes evaluated only on labeled data do not necessarily mean the attributes are informative on the whole partially labeled data, thus resulting in poor performance. Surprisingly, the performance of the selected supervised methods with only labeled data seems rather unstable on some data sets such as "libras" and "wine", where the methods with higher label rates result in worse performance. This inconsistency could be attributed to the scale of initially labeled data. Generally, the smaller the number of training examples, the lower the quality of the reduct and thus the less stable the performance. Laplacian score (LS) is an effective unsupervised attribute reduction method so that all examples in partially labeled data can be utilized. On data sets "gesture1" and "gesture2", FS gains slightly better performance over the proposed method. But on other data sets, the performance of FS is much worse than that of the proposed methods and even signficantly lower than that of the selected supervised method with only labeled data. Different from the supervised and unsupervised methods, the proposed method (Ours) capitalizes on both labeled and unlabeled data, and the unlabeled data are labeled with proxy labels, which are determined by carefully considering the prior information of the whole data and the class distribution of initially labeled data. Moreover, the proposed measure for attribute reduction integrates conditional entropy with information granularity, and the reduct with more discriminant ability could thus be yielded. Over all data sets, the proposed method is maximally improved over GCE-L, FS, and LS by 32.31% ("gesture2" under a label rate $\alpha = 1\%$), 32.80% ("gesture2" under a label rate $\alpha = 1\%$), and 14.47% ("vehicle" under a label rate $\alpha = 30\%$), respectively, when using KNN classifier, and maximally improved over GCE-L, FS, and LS by 22.30% ("krvskp" under a label rate $\alpha = 1\%$), 29.10% ("landsat" under a label rate $\alpha = 5\%$), and 18.11% ("kdd" under a label rate $\alpha = 25\%$), respectively, when using KNN classifier. These results illustrates the effectiveness of the proposed method for partilly labeled data.

Table 5: Settings for all selected methods.

| Method | Attribute evalution | Attribute subset |
| --- | --- | --- |
| Granular conditional entropy with labeled data only (GCE-L) | $Sig(a, P, D)$ | Attribute reduction |
| Fisher score with labeled data only (FS) | $F_{score}(a) = S_B(a)/S_W(a)$ | Top $k$ attbitues |
| Laplacian score with all data without labels (LS) | $L_{score}(a) = \sum_{ij}(a(i)-a(j))^2 S_{ij}/Var(a)$ | Top $k$ attbitues |
| Granular conditional entropy with labeled data and unlabeled data with proxy label (Ours) | $Sig(a, P, D)$ | Attribute reduction |
| Granular conditional entropy with all data labeled (GT) | $Sig(a, P, D)$ | Attribute reduction |
| Data without attribute reduction(Raw) | - | - |

It is worth mentioning that, on data sets "cardio", "frog", and "vehicle", the proposed method outperforms the performance of the original data set without attribute reduction (Raw). This is probably a consequence of attribute reduction and felicitously confirms the fact that attribute reduction could alleviate over-fitting and improve performance. Additionally, on data sets "gesture1", "musk", and "vehicle", the proposed method is improved over the supervised method on the whole data set with true labels (GT), i.e. the data set with a label rate $\alpha = 100\%$, by 1.89% (KNN under a label rate $\alpha = 30\%$), 4.05% (SVM under a label rate $\alpha = 25\%$), and 5.41% (KNN under a label rate $\alpha = 25\%$), respectively. These results may be due to the proposed method selects more informative and discriminative attributes after balancing data set through the labelling strategy for unlabeled data. These findings further demonstrate the proposed method has considerable potential for partially labeled data.

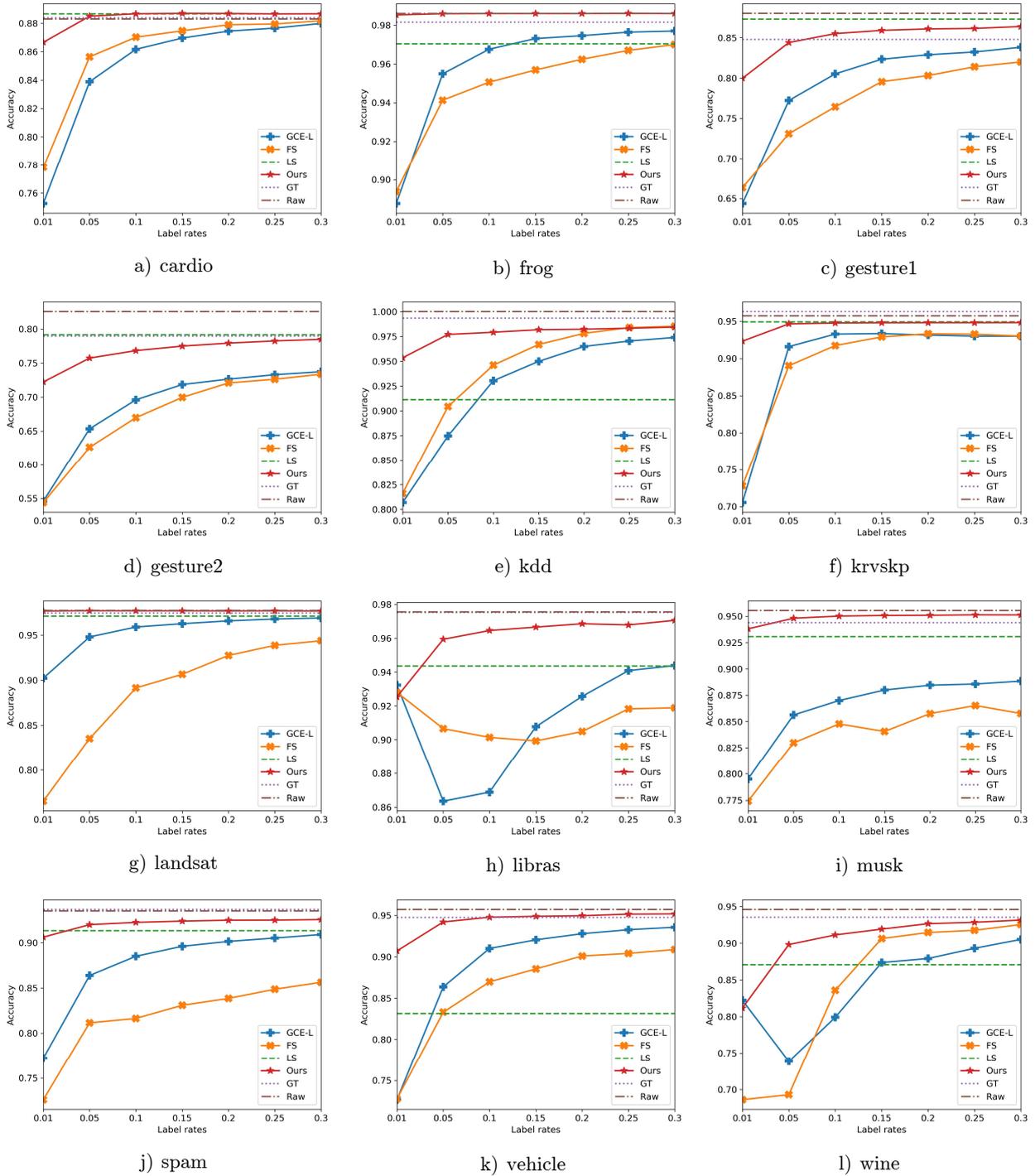

Figure 4. The performance of the selected methods under different label rates (KNN classifier).

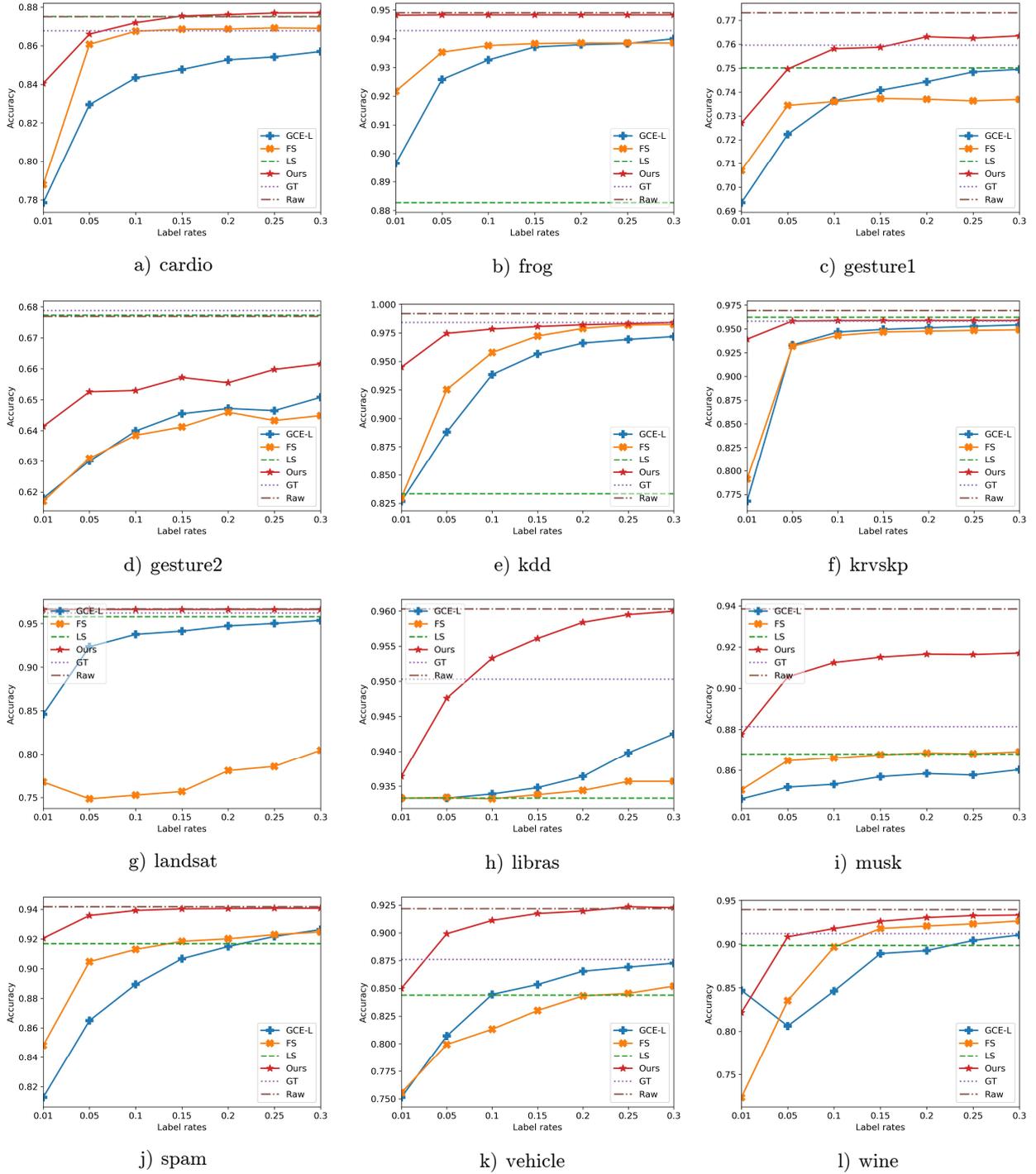

Figure 5. The performance of the selected methods under different label rates (SVM classifier).

## 5. Conclusions

In many real-world tasks, labelling a large number of data is exceptionally costly and practically infeasible so that the data available usually comes with only a few labeled data but a large amount of unlabeled data. In this paper, we propose a simple yet effective strategy to generate the proxy labels for unlabeled data, which not only incorporates the prior knowledge about the whole data, and also considers the class distribution of initially labeled data. To gain the high-quality reduct of partially labeled data with proxy labels, we integrate information granularity into conditional entropy and develop a novel granular conditional entropy, which is theoretically proved to be a monotonic attribute reduction measure. Moreover, a heuristic algorithm based on the proposed granular conditional entropy is designed to quickly induce the optimal reduct of partially labeled data. The experimental results on several benchmark data sets show that the proposed method is effective in dealing with partially labeled data and even performs better than the supervised method on the whole data with true labels. It should be noted that, to deal with numerical attributes, a discretization pre-processing process is involved, an extended method that could directly handle both categorical and numerical attributes is thus expected. Another possible direction is to explore an iterative labelling strategy with the technique of fuzzy clustering [48] to further improve the quality of proxy labels.

## 6. Acknowledgement

The work was supported in part by the National Natural Science Foundation of China (Nos.61806127, 62076164, 62002233), in part by Shenzhen Institute of Artificial Intelligence and Robotics for Society, and in part by the Bureau of Education of Foshan (Nos.2019XJZZ05).

**Appendix**

Proof of Proposition 2.

$\Delta GH = GH(D|P) - GH(D|Q)$

$= P(X_i)^2 \sum_{k=1}^{|U/D|} P(Y_k|X_i)P(Y_k|X_i) + P(X_j)^2 \sum_{k=1}^{|U/D|} P(Y_k|X_j)P(Y_k|X_j)$

$- P(X_{ij})^2 \sum_{k=1}^{|U/D|} P(Y_k|X_{ij})P(Y_k|X_{ij})$

$= \frac{1}{|U|^2} \sum_{k=1}^{|U/D|} \left( |X_i|^2 \frac{|X_i \cap Y_k|}{|X_i|} \log \frac{|X_i \cap Y_k|}{|X_i|} + |X_j|^2 \frac{|X_j \cap Y_k|}{|X_j|} \log \frac{|X_j \cap Y_k|}{|X_j|} \right.$

$\left. - |X_{ij}|^2 \frac{|X_{ij} \cap Y_k|}{|X_{ij}|} \log \frac{|X_{ij} \cap Y_k|}{|X_{ij}|} \right)$

Let $\theta_i = |X_i \cap Y_k|/|X_i|, \theta_j = |X_j \cap Y_k|/|X_j|, \theta_{ij} = |X_{ij} \cap Y_k|/|X_{ij}|$. We have

$\Delta GH = \frac{1}{|U|^2} \sum_{k=1}^{|U/D|} (|X_i|^2 \theta_i \log \theta_i + |X_j|^2 \theta_j \log \theta_j - |X_{ij}|^2 \theta_{ij} \log \theta_{ij})$

and $|X_{ij}|\theta_{ij} = |X_i|\theta_i + |X_j|\theta_j$ for any decision $Y_k$.

Let $f_k = |X_i|^2 \theta_i \log \theta_i + |X_j|^2 \theta_j \log \theta_j - |X_{ij}|^2 \theta_{ij} \log \theta_{ij}$. Then, we have

$\Delta GH = \frac{1}{|U|^2} \sum_{k=1}^{|U/D|} f_k.$

For any $k$, we have

$f_k = |X_i|^2 \theta_i \log \theta_i + |X_j|^2 \theta_j \log \theta_j - |X_{ij}|^2 \theta_{ij} \log \theta_{ij}$

$= |X_i|^2 \theta_i \log \theta_i + |X_j|^2 \theta_j \log \theta_j - |X_{ij}|(|X_i|\theta_i + |X_j|\theta_j) \log \left( \frac{|X_i|\theta_i + |X_j|\theta_j}{|X_{ij}|} \right)$

$$\geq |X_i|^2\theta_i\log\theta_i + |X_j|^2\theta_j\log\theta_j - (|X_i|^2\theta_i + |X_j|^2\theta_j)\log\left(\frac{|X_i|\theta_i + |X_j|\theta_j}{|X_{ij}|}\right)$$

$$= |X_i|^2\theta_i\left(\log\theta_i - \log\left(\frac{|X_i|\theta_i + |X_j|\theta_j}{|X_{ij}|}\right)\right)$$

$$+ |X_j|^2\theta_j\left(\log\theta_j - \log\left(\frac{|X_i|\theta_i + |X_j|\theta_j}{|X_{ij}|}\right)\right)$$

$$= |X_i|^2\theta_i\log\left(\frac{|X_i|\theta_i + |X_j|\theta_i}{|X_i|\theta_i + |X_j|\theta_j}\right) + |X_j|^2\theta_j\log\left(\frac{|X_i|\theta_j + |X_j|\theta_j}{|X_i|\theta_i + |X_j|\theta_j}\right)$$

Let the right side of the above formula be $f'_k$ and $m = |X_i|$, $n = |X_j|$, $\mu = |X_i|\theta_i$, $\nu = |X_j|\theta_j$, and $\lambda = \theta_i/\theta_j$. We have

$$f'_k(\mu, \nu, \lambda) = m\mu\log\left(\frac{\mu + \lambda\nu}{\mu + \nu}\right) + n\nu\log\left(\frac{\mu/\lambda + \nu}{\mu + \nu}\right).$$

$f'_k(\mu, \nu, \lambda)$ is an explicit function of the variables $\mu, \nu$, and $\lambda$. The partial derivative of $f'_k(\mu, \nu, \lambda)$ with respect to the variable $\lambda$ is

$$\frac{\partial f'_k(\mu, \nu, \lambda)}{\partial \lambda} = m\mu\left(\frac{\mu + \nu}{\mu + \lambda\nu}\right)\nu\log 2 - n\nu\left(\frac{\mu + \nu}{\mu/\lambda + \nu}\right)\frac{\mu}{\lambda^2}\log 2$$

$$= \log 2\left(\mu\nu(\mu + \nu)\left(\frac{\lambda m - n}{\lambda(\mu + \nu\lambda)}\right)\right)\begin{cases} < 0, & 0 < \lambda < n/m \\ = 0, & \lambda = n/m \\ > 0, & \lambda > n/m \end{cases}.$$

$f'_k(\mu, \nu, \lambda)$ arrives at the minima when $\lambda = n/m$. Namely, $|X_i|\theta_i = |X_j|\theta_j$. In this case, $f'_k(\mu, \nu, \lambda) = 0$ and $\Delta GH = 0$. Thus, $\Delta GH \geq 0$ holds for every possible case. The proposition is proved.

## References

[1] C. Bishop, Pattern Recognition and Machine Learning, Springer, New York, NY, USA, 2007.

[2] V. Bolón-Canedo, N. Sánchez-Maroño, A. Alonso-Betanzos, Recent advances and emerging challenges of feature selection in the context of big data, Knowl.-Based Syst., 86 (2015) 33-45.

[3] O. Chapelle, B. Scholkopf, A. Zien, Semi-supervised learning, MIT Press, Cambridge, MA, USA, 2006.

[4] J.H. Dai, Q.H. Hu, J.H. Zhang, H. Hu, N.G. Zheng, Attribute selection for partially labeled categorical data by rough set approach, IEEE Trans. Cybern., 47 (2017) 2460-2471.

[5] C. Gao, Z.H. Lai, J. Zhou, C.R. Zhao, D.Q. Miao, Maximum decision entropy-based attribute reduction in decision-theoretic rough set model, Knowl.-Based Syst., 143 (2018) 179-191.

[6] C. Gao, Z.H. Lai, J. Zhou, J.J. Wen, W.K. Wong, Granular maximum decision entropy-based monotonic uncertainty measure for attribute reduction, Int. J. Approx. Reason., 104 (2019) 9-24.

[7] I. Guyon, A. Elisseeff, An introduction to variable and feature selection, J. Mach. Learn. Res., 3 (2003) 1157-1182.

[8] X.F. He, D. Cai, P. Niyogi, Laplacian score for feature selection, in: Proceedings of the 18th Advances in Neural Information Processing Systems, Vancouver, BC, Canada, 2005, pp. 507-514.

[9] X.H. Hu, N. Cercone, Learning in relational databases: A rough set approach, Comput. Intell., 11 (1995) 323-338.

[10] R. Jensen, S. Vluymans, N. Mac Parthalain, C. Cornelis, Y. Saeys, Semi-supervised fuzzy-rough feature selection, in: Proceedings of the 15th International Conference on Rough Sets, Fuzzy Sets, Data Mining, and Granular Computing, Tianjin, China, 2015, pp. 185-195.

[11] X.Y. Jia, L. Shang, B. Zhou, Y.Y. Yao, Generalized attribute reduct in rough set theory, Knowl.-Based Syst., 91 (2016) 204-218.

[12] X.Y. Jia, W.H. Liao, Z.M. Tang, L. Shang, Minimum cost attribute reduction in decision-theoretic rough set models, Inf. Sci., 219 (2013) 151-167.
[13] C.C. Kuo, H.L. Shieh, A semi-supervised learning algorithm for data classification, Int. J. Pattern Recognit Artif Intell., 29 (2015) 1551007.
[14] J.Y. Liang, Z.Z. Shi, The information entropy, rough entropy and knowledge granulation in rough set theory, Int. J. Uncertain. Fuzziness Knowl.-Based Syst., 12 (2004) 37-46.
[15] B.Y. Li, J.M. Xiao, X.H. Wang, Feature selection for partially labeled data based on neighborhood granulation measures, IEEE Access, 7 (2019) 37238-37250.
[16] B.Z. Li, Z.H. Wei, D.Q. Miao, N. Zhang, W. Shen, C. Gong, H.Y. Zhang, L.J. Sun, Improved general attribute reduction algorithms, Inf. Sci., 536 (2020) 298-316.
[17] J.D. Li, K.W. Cheng, S.H. Wang, F. Morstatter, R.P. Trevino, J.L. Tang, H. Liu, Feature selection: A data perspective, ACM Comput. Surv., 50 (2018) 1-45.
[18] P. Lingras, M. Chen, D.Q. Miao, Semi-supervised rough cost/benefit decisions, Fundam. Informaticae, 94 (2009) 233-244.
[19] K.Y. Liu, E.C.C. Tsang, J.J. Song, H.L. Yu, X.J. Chen, X.B Yang, Neighborhood attribute reduction approach to partially labeled data, Granul. Comput., 5 (2020) 239-250.
[20] K.Y. Liu, X.B. Yang, H.L. Yu, J.X. Mi, P.X. Wang, X.J. Chen, Rough set based semi-supervised feature selection via ensemble selector, Knowl.-Based Syst., 165 (2019) 282-296.
[21] Y. Li, T. Li, H. Liu, Recent advances in feature selection and its applications, Knowl. Inf. Syst., 53 (2017) 551-577.
[22] D.Q. Miao, C. Gao, N. Zhang, Z.F. Zhang, Diverse reduct subspaces based co-training for partially labeled data, Int. J. Approx. Reason., 52 (2011) 1103-1117.
[23] D.Q. Miao, G.R. Hu, A heuristic algorithm for reduction of knowledge, J. Comput. Res. Dev., 36 (1999) 681-684 (in Chinese).
[24] D.Q. Miao, Y. Zhao, Y.Y. Yao, H.X. Li, F.F. Xu, Relative reducts in consistent and inconsistent decision tables of the Pawlak rough set model, Inf. Sci., 179 (2009) 4140-4150.
[25] F. Min, F.L. Liu, L.Y. Wen, Z.H. Zhang, Tri-partition cost-sensitive active learning through kNN, Soft Comput., 23 (2019) 1557-1572.
[26] S. Miyamoto, S. Takumi, Hierarchical clustering using transitive closure and semi-supervised classification based on fuzzy rough approximation, in: Proceedings of 2012 IEEE International Conference on Granular Computing, 2012, pp. 359-364.
[27] N.M. Parthalain, R. Jensen, Fuzzy-rough set based semi-supervised learning, in: Proceedings of 2011 IEEE International Conference on Fuzzy Systems, Taipei, Taiwan, 2011, pp. 2465-2472.
[28] Z. Pawlak, Rough sets, Int. J. Comput. Inf. Sci., 11 (1982) 341-356.
[29] Z. Pawlak, Rough sets: Theoretical aspects of reasoning about data, Kluwer Academic Publishers, Dordrecht, Netherlands, 1991.
[30] Z. Pawlak, S.K.M. Wong, W. Ziarko, Rough sets: Probabilistic versus deterministic approach, Int. J. Man-Mach. Stud., 29 (1988) 81-95.
[31] J. Qian, C.L. Liu, D.Q. Miao, X.D. Yue, Sequential three-way decisions via multi-granularity, Inf. Sci., 507 (2020) 606-629.
[32] Y.H. Qian, J.Y. Liang, W. Pedrycz, C.Y. Dang, Positive approximation: An accelerator for attribute reduction in rough set theory, Artif. Intell., 174 (2010) 597-618.
[33] R. Sheikhpour, M.A. Sarram, S. Gharaghani, M.A.Z. Chahooki, A Survey on semi-supervised feature selection methods, Pattern Recognit., 64 (2017) 141-158.
[34] L. Shi, X.M. Ma, L. Xi, Q.G. Duan, J.Y. Zhao, Rough set and ensemble learning based semi-supervised algorithm for text classification, Expert Syst. Appl., 38 (2011) 6300-6306.
[35] K. Thangavel, A. Pethalakshmi, Dimensionality reduction based on rough set theory: a review, Appl. Soft Comput., 9 (2009) 1-12.
[36] I. Triguero, S. García, F. Herrera, Self-labeled techniques for semi-supervised learning: Taxonomy, software and empirical study, Knowl. Inf. Syst., 42 (2015) 245-284.
[37] R. Wang, D.G Chen, S. Kwong, Fuzzy-rough-set-based active learning, IEEE Trans. Fuzzy Syst., 22 (2014) 1699-1704.
[38] Z.X. Xue, Y.L. Shang, A.F. Feng, Semi-supervised outlier detection based on fuzzy rough C-means clustering, Math. Comput. Simul., 80 (2010) 1911-1921.

[39] Y.Y. Yao, Three-way decisions with probabilistic rough sets, Inf. Sci., 180 (2010) 341-353.

[40] Y.Y. Yao, The superiority of three-way decisions in probabilistic rough set models, Inf. Sci., 181 (2011) 1080-1096.

[41] Y.Y. Yao, Tri-level thinking: models of three-way decision, Int. J. Mach. Learn. Cybern., 11 (2020) 947-959.

[42] Y.Y. Yao, Y. Zhao, Attribute reduction in decision-theoretic rough set models, Inf. Sci., 178 (2008) 3356-3373.

[43] Y.Y. Yao, Y. Zhao, Discernibility matrix simplification for constructing attribute reducts, Inf. Sci., 179 (2009) 867-882.

[44] X.D. Yue, J. Zhou, Y.Y. Yao, D.Q. Miao, Shadowed neighborhoods based on fuzzy rough transformation for three-way classification, IEEE Trans. Fuzzy Syst., 28 (2020) 978-991.

[45] Q.H. Zhang, Q. Xie, G.Y. Wang, A survey on rough set theory and its applications, CAAI Trans. Intell. Technol., 1 (2016) 323-333.

[46] W. Zhang, D.Q. Miao, C. Gao, X.D. Yue, Co-training based attribute reduction for partially labeled data, in: Proceedings of the 9th International Conference on Rough Sets and Knowledge Technology, Shanghai, China, 2014, pp. 77-88.

[47] J. Zhou, C. Gao, W. Pedrycz, Z.H. Lai, X.D. Yue, Constrained shadowed sets and fast optimization algorithm, Int. J. Intell. Syst., 34 (2019) 2655-2675.

[48] J. Zhou, W. Pedrycz, X.D. Yue, C. Gao, Z.H. Lai, J. Wan, Projected fuzzy C-means clustering with locality preservation, Pattern Recognit., https://doi.org/10.1016/j.patcog.2020.107748.

[49] Z.H. Zhou, A brief introduction to weakly supervised learning, National Science Review, 5 (2018) 48-57.

[50] X.J. Zhu, and A.B. Goldberg, Introduction to semi-supervised learning, Morgan & Claypool Publishers, Cambridge, MA, USA, 2009.